\documentclass{article}

\usepackage[utf8]{inputenc}
\usepackage[T1]{fontenc}
\usepackage{amsmath,amssymb,amsfonts}
\usepackage{graphicx}
\usepackage{booktabs}
\usepackage{multirow}
\usepackage{hyperref}
\usepackage{xcolor}
\usepackage{algorithm}
\usepackage{algorithmic}
\usepackage[margin=1in]{geometry}
\usepackage{natbib}
\usepackage{subcaption}

\graphicspath{{./}}

\title{Self-Monitoring Benefits from Structural Integration:\\
Lessons from Metacognition in\\
Continuous-Time Multi-Timescale Agents}

\author{
Ying Xie\\
Kennesaw State University
}

\date{}

\begin{document}
\maketitle

\begin{abstract}
Self-monitoring capabilities---metacognition, self-prediction, and subjective duration---are often proposed as useful additions to reinforcement learning agents. But do they actually help? We investigate this question in a continuous-time multi-timescale agent operating in predator-prey survival environments of varying complexity, including a 2D partially observable variant. We first show that three self-monitoring modules, implemented as auxiliary-loss add-ons to a multi-timescale cortical hierarchy, provide \emph{no statistically significant benefit} across 20 random seeds, 1D and 2D predator-prey environments with standard and non-stationary variants, and training horizons up to 50{,}000 steps. Diagnosing the failure, we find the modules collapse to near-constant outputs (confidence std $< 0.006$, attention allocation std $< 0.011$) and the subjective duration mechanism shifts the discount factor by less than $0.03\%$. Policy sensitivity analysis confirms the agent's decisions are unaffected by module outputs in this design. We then show that \emph{structurally integrating} the module outputs---using confidence to gate exploration, surprise to trigger workspace broadcasts, and self-model predictions as policy input---produces a medium-large improvement over the add-on approach (Cohen's $d = 0.62$, $p = 0.06$, paired) in a non-stationary environment. Component-wise ablations reveal that the TSM-to-policy pathway contributes most of this gain. However, structural integration does not significantly outperform a baseline with no self-monitoring ($d = 0.15$, $p = 0.67$), and a parameter-matched control without modules performs comparably, so the benefit may lie in recovering from the trend-level harm of ignored modules rather than in self-monitoring content. The architectural implication is that self-monitoring should sit on the decision pathway, not beside it.
\end{abstract}

\section{Introduction}

A predator moves at a different timescale than the seasons change. An animal surviving in the wild must simultaneously track fast threats (a lunging predator), medium-term opportunities (a ripening fruit patch), and slow environmental patterns (the migration of prey herds). The neural architectures that support this kind of multi-timescale integration are well-studied~\citep{murray2014hierarchy, hasson2015hierarchical}. This raises the question of whether agents that process information across timescales should also \emph{monitor their own processing}---maintaining confidence estimates, predicting their own future states, and adjusting temporal horizons based on what is happening internally.

Theories of consciousness---Global Workspace Theory~\citep{baars1988cognitive, dehaene2011experimental}, Integrated Information Theory~\citep{tononi2016integrated}, Higher-Order Theories~\citep{rosenthal2006consciousness}, and the Attention Schema Theory~\citep{graziano2013consciousness}---each propose mechanisms by which biological systems achieve self-monitoring, temporal integration, and adaptive regulation. Recent work has begun connecting these ideas to computational architectures: \citet{goyal2022coordination} implemented a shared workspace for modular networks, and \citet{juliani2022deep} explored theoretical links between conscious function and general intelligence in agents. The implicit assumption in this line of work is that self-monitoring capabilities, once added to an agent, will improve performance. We test this assumption.

We build a continuous-time agent with a multi-timescale cortical hierarchy and equip it with three self-monitoring modules inspired by theories of consciousness:

\begin{enumerate}
    \item \textbf{Metacognition}: confidence estimates, surprise detection, and attention allocation across processing timescales.
    \item \textbf{Temporal self-model}: prediction of the agent's own future internal states, creating a form of self-knowledge.
    \item \textbf{Subjective duration}: a learned ``felt time'' signal that modulates the discount factor.
\end{enumerate}

Our investigation proceeds in three stages:

\begin{enumerate}
    \item \textbf{Null result.} When implemented as auxiliary-loss add-ons (the standard approach), the three modules provide no statistically significant benefit. Across 20 seeds, 1D and 2D predator-prey environments (standard and non-stationary variants), and training horizons up to 50{,}000 steps, the full model never significantly outperforms a baseline with no self-monitoring. A parameter-matched control and a random-auxiliary-loss control perform equally well.

    \item \textbf{Diagnosis.} The modules fail because their outputs collapse to near-constants that the agent learns to ignore. Confidence varies by less than $0.006$ std across all seeds; attention allocation by less than $0.011$, even around deaths. Subjective duration shifts the discount by under $0.03\%$. Policy sensitivity analysis confirms that perturbing module outputs produces negligible change in the agent's action distribution.

    \item \textbf{Structural fix.} Replacing the add-on design with \emph{structural integration}---routing confidence into exploration gating, surprise into workspace triggering, and self-model predictions into the policy head---produces a substantial improvement over the add-on approach (Cohen's $d = 0.62$) in a non-stationary environment. Component-wise ablations show that the TSM-to-policy pathway contributes most to this gain. However, structurally integrated self-monitoring does not significantly outperform a no-self-monitoring baseline, and a param-matched control without modules performs comparably, so we cannot attribute the gain to self-monitoring content rather than additional capacity.
\end{enumerate}

The contribution is a practical architectural lesson: in the environments we study, self-monitoring for RL agents is more effective when the decision pathway \emph{depends on} the monitoring signals, and fails when the signals are attached as optional features via auxiliary losses. We support this claim with controlled experiments, parameter-matched and random-auxiliary baselines, component-wise ablations, and policy sensitivity analysis. We do not claim that these modules produce phenomenal consciousness~\citep{chalmers1995facing, block1995confusion}; our contribution is about engineering self-monitoring into agents in a way that avoids the auxiliary-loss trap. Generalization to larger-scale agents and more complex environments is untested.

\section{Related Work}

\paragraph{Multi-timescale recurrent networks.}
The idea that neural systems should operate at multiple temporal resolutions is well-established in neuroscience~\citep{murray2014hierarchy} and machine learning~\citep{el1995hierarchical, chung2017hierarchical}. Liquid time-constant networks~\citep{hasani2021liquid} developed input-dependent time constants for continuous-time neural ODEs. Neural circuit policies~\citep{lechner2020neural} demonstrated that such architectures can learn interpretable control policies. Our base architecture builds on this line of work, extending liquid cells with Hebbian plasticity and exponential moving average memory.

\paragraph{Global workspace models.}
Global Workspace Theory~\citep{baars1988cognitive} proposes that conscious processing involves broadcasting information into a shared workspace. \citet{goyal2022coordination} implemented this as a bottleneck attention mechanism between modular networks. \citet{juliani2022deep} explored agents with workspace-inspired architectures. These works add workspace mechanisms as structural components; we examine when such components actually contribute to task performance.

\paragraph{Metacognition and auxiliary losses in RL.}
Self-monitoring in reinforcement learning has been approached through uncertainty-driven exploration~\citep{osband2016deep}, confidence calibration in neural networks~\citep{guo2017calibration}, and meta-learning~\citep{wang2016learning}. \citet{cleeremans2020learning} argued that metacognition should be understood as the system learning to redescribe its own representations. A common pattern in these approaches is to add self-monitoring as an \emph{auxiliary loss}: the module is trained to predict some internal quantity, and the resulting gradients improve the shared representation. Our results show that this approach can fail completely when the module outputs are not used for decision-making.

\paragraph{Self-models and predictive processing.}
Predictive processing frameworks~\citep{clark2013whatever, friston2010free} emphasize that brains continuously predict their own sensory inputs. \citet{hafner2020dream} showed that world models can support planning in RL. Our temporal self-model predicts the agent's own future \emph{internal} states rather than external observations, closer to the interoceptive self-models proposed by~\citet{seth2012interoceptive}. We find that self-prediction only helps when the predictions feed into the policy, not when they are trained solely as an auxiliary objective.

\paragraph{Subjective time perception.}
The psychological literature on time perception~\citep{eagleman2008human, wittmann2009inner} suggests that subjective temporal experience varies with event density and threat. Computational models of interval timing~\citep{matell2004cortico} typically focus on psychophysical data. Our subjective duration module attempts to use felt time as a learning signal by modulating the discount factor, but we find the effect is negligibly small in practice ($< 0.03\%$ change in $\gamma$).

\paragraph{Negative results in consciousness-inspired AI.}
Recent work has examined whether AI systems satisfy proposed criteria for consciousness~\citep{butlin2023consciousness} and developed theoretical frameworks connecting consciousness to computation~\citep{seth2021being}. We ask a different question: whether \emph{functional} analogs of consciousness-associated capabilities actually improve agent performance. For add-on implementations, the answer is no.

\section{Method}

\subsection{Environment}

The agent lives in a one-dimensional toroidal world of size $W=100$ with wraparound boundaries. The world contains $N_f=5$ food sources and $N_p=2$ predators. Food sources are stationary; predators chase the agent with speed 0.6 (the agent moves at speed 1.0) plus Gaussian noise. Time advances in variable steps $\Delta t$ drawn from a log-normal distribution with mean 1.0 and standard deviation 0.3, clamped to $[0.5, 2.0]$.

Eating food (within radius 2.0) gives reward $+1$, restores 0.2 energy, and respawns the food at a random location. Getting caught by a predator (within radius 1.5) gives reward $-5$ and teleports the agent to a random position. The agent has an energy level (clamped to $[0, 2]$) that decays at $0.001 \cdot \Delta t$ per step. A periodic danger wave sweeps across the world with period 200 time units, active for 30\% of each cycle; the danger zone has spatial width 15. Being inside the danger zone incurs a continuous penalty of $-2 \cdot \Delta t$.

The observation vector ($d=30$) consists of relative positions and two type flags (is-food, is-predator) for each of the 8 nearest entity slots (with 7 real entities and one zero-padded slot), the agent's normalized position, velocity, and energy level, and three danger-wave features (phase, active flag, relative distance to danger center).

\subsection{Base Architecture: Multi-Timescale Cortical Hierarchy}

Figure~\ref{fig:architecture} shows the full architecture. The agent processes observations through three \emph{Plastic Cortical Cells} arranged in a feedforward hierarchy. Each cell extends a liquid time-constant neural ODE~\citep{hasani2021liquid} with two forms of plasticity:

\begin{figure}[t]
\centering
\includegraphics[width=0.85\textwidth]{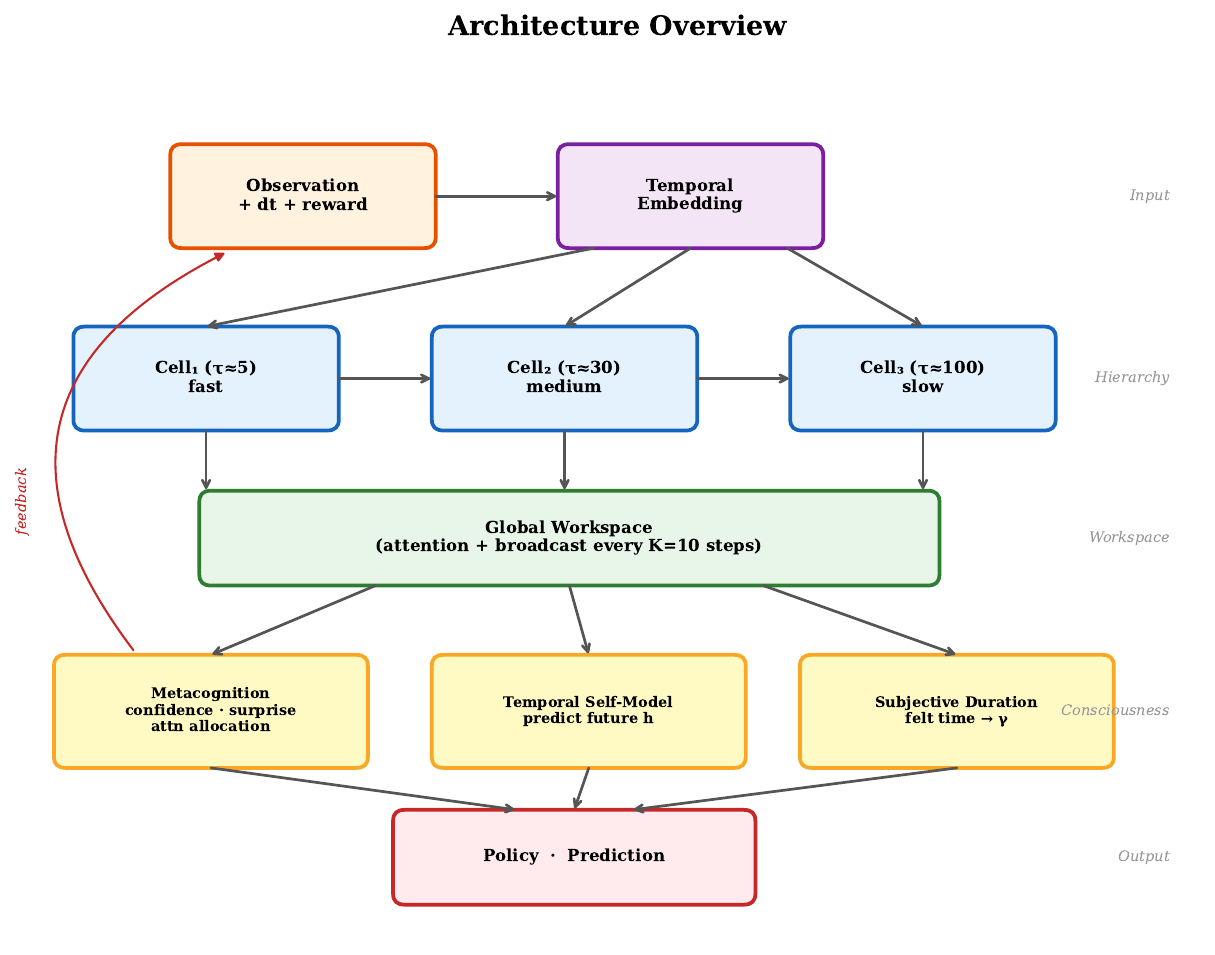}
\caption{Architecture overview. Observations pass through a temporal embedding and three cortical cells with increasing time constants. A global workspace broadcasts across levels. Three self-monitoring modules (bottom) monitor the hierarchy. In the add-on design, module outputs are fed back as input features. In the structural variant, they sit on the decision critical path: confidence gates exploration, surprise triggers workspace broadcasts, and self-model predictions enter the policy head.}
\label{fig:architecture}
\end{figure}

\paragraph{Liquid dynamics.} Each cell computes:
\begin{align}
    \alpha &= \frac{\Delta t}{\tau + \Delta t} \\
    x_{\text{new}} &= \tanh\!\big(W_{\text{in}} \cdot \text{input} + W_{\text{rec}} \cdot h + T \cdot h + W_{\text{mem}} \cdot m\big) \\
    h' &= (1 - \alpha) \cdot h + \alpha \cdot x_{\text{new}}
\end{align}
where $\tau$ is a learnable time constant (stored in log-space), $h$ is the hidden state, and $\alpha$ controls how much new information is incorporated at each step. Fast cells ($\tau \approx 5$) update quickly; slow cells ($\tau \approx 100$) retain information across many steps.

\paragraph{Hebbian trace.} A rank-one Hebbian learning rule maintains a fast associative trace $T$:
\begin{equation}
    T' = \lambda_d \cdot T + \lambda_h \cdot x_{\text{new}} \cdot h^\top
\end{equation}
with decay $\lambda_d = 0.95$ and learning rate $\lambda_h = 0.01$. The trace allows rapid one-shot association between inputs and hidden states without gradient-based learning.

\paragraph{EMA memory.} An exponential moving average $m' = 0.99 \cdot m + 0.01 \cdot h'$ provides a slowly changing context signal to each cell.

The three cells have hidden dimension $d_h = 32$ and are initialized with $\tau_1 = 5$ (fast), $\tau_2 = 30$ (medium), $\tau_3 = 100$ (slow). Cell 1 receives the projected observation; each subsequent cell receives the layer-normalized output of the previous cell.

\paragraph{Interleaved global workspace.} Every $K=10$ steps, a transformer encoder with 2 attention heads collects the last $L=5$ hidden states from all three levels, runs self-attention across the 15 tokens (5 per level), and broadcasts information back via gated addition. The gates are initialized with bias $-2.0$ (near zero), so the workspace starts inert and learns when cross-level communication helps.

\subsection{Metacognition Module}

The metacognition module receives the concatenated hidden states from all three levels ($[h_1, h_2, h_3] \in \mathbb{R}^{96}$) and produces two groups of outputs.

\textbf{Confidence, uncertainty, and attention allocation} are produced by a two-layer MLP ($96 \to 32 \to 6$). The first output (sigmoid) gives a scalar confidence estimate; the second (softplus) gives an uncertainty estimate. Three further outputs pass through a softmax to produce an attention allocation over cortical levels, modulating the workspace broadcast gates. When the metacognition module assigns high weight to the slow level, the workspace broadcasts more from level 3.

\textbf{Surprise} is computed separately by maintaining a low-rank prediction of the next step's hidden states. At each step, the module predicts $\hat{h}_i^{t+1}$ for each level via down-projection ($32 \to 8$) followed by up-projection ($8 \to 32$). At the next step, surprise is the MSE between predicted and actual states, averaged across levels. This surprise signal is fed back as input to the agent on subsequent steps.

The full metacognition signal concatenates both groups: $[\text{confidence}, \text{uncertainty}, \text{attn\_alloc}(3), \text{surprise}] \in \mathbb{R}^6$.

The auxiliary loss for metacognition is the surprise prediction error itself, weighted by $\lambda_m = 0.05$.

\subsection{Temporal Self-Model (TSM)}

The temporal self-model (TSM) predicts the agent's own hidden states $H=5$ steps into the future. An encoder ($96 \to 16$) compresses the current state, a horizon embedding modulates the compressed representation, and a decoder ($16 \to 96$) produces the predicted future states for all three levels:

\begin{align}
    z &= \text{ReLU}(W_{\text{enc}} \cdot [h_1, h_2, h_3]) \\
    z_{\text{mod}} &= z \odot \sigma(W_{\text{hor}} \cdot H) \\
    [\hat{h}_1^{t+H}, \hat{h}_2^{t+H}, \hat{h}_3^{t+H}] &= W_{\text{dec}} \cdot z_{\text{mod}}
\end{align}

A buffer of the last $H{+}5{=}10$ hidden-state snapshots stores past states. When the buffer contains an entry from $H$ steps ago, the model re-runs its prediction from those stored states and compares with the current actual states. This delayed comparison is the self-prediction loss ($\lambda_s = 0.05$).

A separate predictability head ($16 \to 3$, sigmoid) learns per-level scores indicating how self-predictable each cortical level is. These scores are not used for decision-making; they emerge as a learned self-description.

\subsection{Subjective Duration Module}

The subjective duration module produces a scalar ``felt duration'' per step, representing the agent's learned sense of how much is happening. It receives 8 features:

\begin{equation}
    \text{features} = [\Delta t,\; \|h_1' - h_1\|,\; \|h_2' - h_2\|,\; \|h_3' - h_3\|,\; \text{surprise},\; |r|,\; \text{danger},\; \text{confidence}]
\end{equation}

A context projection ($96 \to 16$) and a feature projection ($8 \to 16$) are concatenated and passed through a two-layer MLP ($32 \to 16 \to 1$). The output passes through softplus plus a floor of 0.1, ensuring positivity.

The felt duration modulates the discount factor during REINFORCE updates:
\begin{equation}
    \gamma_{\text{eff}} = \gamma^{1 + \lambda_\gamma \cdot (d_{\text{felt}} - 1)}
\end{equation}
where $\gamma = 0.99$, $\lambda_\gamma = 0.1$, and $\gamma_{\text{eff}}$ is clamped to $[0.9, 0.999]$. When felt duration is high (dense events), $\gamma_{\text{eff}}$ decreases, making the agent weight immediate rewards more heavily. When felt duration is low (calm period), $\gamma_{\text{eff}}$ increases, encouraging longer-term planning.

The module has no direct loss---it is trained entirely through the effect of $\gamma_{\text{eff}}$ on the policy gradient.

\subsection{Training}

The agent chooses among three discrete actions (move left, stay, move right) via a two-layer policy head ($d_h \to d_h \to 3$) with softmax output, trained with REINFORCE and an EMA reward baseline (decay 0.99). An observation prediction head (linear, $d_h \to d_{\text{obs}}$) applied to the deepest cortical level predicts the next observation; its MSE loss $\mathcal{L}_{\text{pred}}$ provides an auxiliary learning signal. The entropy bonus $\mathcal{L}_{\text{entropy}}$ is estimated as the negative mean log-probability of the sampled actions within each BPTT window (a single-sample approximation to the policy entropy). The optimizer is Adam with learning rate $3 \times 10^{-4}$. The total loss combines:
\begin{equation}
    \mathcal{L} = \mathcal{L}_{\text{policy}} + \lambda_e \mathcal{L}_{\text{entropy}} + \lambda_p \mathcal{L}_{\text{pred}} + \lambda_m \mathcal{L}_{\text{surprise}} + \lambda_s \mathcal{L}_{\text{self-pred}}
\end{equation}
with $\lambda_e = 0.01$, $\lambda_p = 0.1$, $\lambda_m = 0.05$, $\lambda_s = 0.05$. Gradients are clipped to norm 1.0. Updates occur every 50 steps (the BPTT window), after which all hidden states are detached.

\subsection{Structural Integration (Fix)}

The add-on design described above feeds module outputs back as input features that the agent \emph{may learn to use}. Our structural integration variant instead places the module outputs on the decision-making critical path, so the agent cannot bypass them:

\paragraph{Confidence $\to$ exploration gating.} Instead of a fixed entropy coefficient $\lambda_e$, we use $\lambda_e \cdot (1.5 - c)$, where $c$ is the mean confidence over the BPTT window. Low confidence increases exploration; high confidence reduces it.

\paragraph{Surprise $\to$ workspace trigger.} Instead of broadcasting every $K{=}10$ steps, the workspace fires when the surprise signal exceeds a learned threshold $\sigma(\theta_s)$. This makes the global workspace reactive to unexpected events rather than periodic.

\paragraph{Self-model predictions $\to$ policy input.} The policy head receives two inputs: (1) a predictability-weighted combination of all three cortical levels $h_w = \sum_i \text{softmax}(5 \cdot p_i) \cdot h_i$ where $p_i$ are the TSM's per-level predictability scores, and (2) the TSM's predicted future state at the deepest level, detached from the TSM's gradient graph. The policy head thus sees both the current state (weighted by self-assessed reliability) and a forecast of the agent's own future internal state.

These changes increase the policy head input dimension from $d_h$ to $2 d_h$ but add no new learnable parameters beyond the surprise threshold scalar.

\subsection{Non-Stationary Environment Variant}

To test conditions where self-monitoring should theoretically help, we also evaluate on a harder variant with:
\begin{itemize}
    \item \textbf{Predator phases}: predators alternate between aggressive (full speed) and passive ($0.2\times$ speed) phases with period 500 time units. The agent must detect which phase it is in.
    \item \textbf{Poison food}: 30\% of food items are poisonous (reward $-2$ instead of $+1$), visually indistinguishable from safe food. The agent must track its own health response.
    \item \textbf{Noisy observations}: Gaussian noise ($\sigma = 0.15$) and random dropout (15\%) are applied to the observation vector.
\end{itemize}

\subsection{2D Partially Observable Variant}

To test whether our findings generalize beyond the 1D setting, we implement a 2D toroidal environment ($[0, W)^2$) with five discrete actions (stay, up, down, left, right). The observation space expands to $d=39$: 8 nearest entities $\times$ 4 features (relative $x$, relative $y$, is-food, is-predator) plus 5 agent-state features and 2 danger features. All other dynamics (food, predators, danger waves) carry over with appropriate 2D distance calculations. When combined with the non-stationary flags (noisy observations, predator phases, poison food), this creates a richer setting where self-monitoring should in principle be more valuable.

\subsection{Component-Wise Structural Ablation}

To identify which structural pathway contributes most to the improvement, we test three single-pathway variants:
\begin{itemize}
    \item \textbf{Confidence only}: confidence gates entropy coefficient, but workspace remains periodic and TSM does not enter the policy head.
    \item \textbf{Surprise only}: surprise triggers reactive workspace broadcasts, but entropy is fixed and TSM does not enter the policy head.
    \item \textbf{TSM only}: self-model predictions and predictability-weighted states enter the policy head, but entropy is fixed and workspace remains periodic.
\end{itemize}

Each variant uses the full set of self-monitoring modules but only structurally integrates one pathway, allowing us to isolate each component's contribution.

\subsection{Policy Sensitivity Analysis}

To directly measure whether module outputs affect decisions, we compute the KL divergence between the agent's policy under normal conditions and under perturbation of individual module outputs. At every 100 steps during evaluation, we perturb each module signal (confidence, surprise, predictability) by a fixed magnitude ($\delta = 0.5$), re-run the policy head, and measure $D_{\text{KL}}(\pi_{\text{base}} \| \pi_{\text{perturbed}})$. High KL indicates the policy is sensitive to the module output; near-zero KL confirms the module is being ignored. This analysis provides direct evidence for the ignore-ability hypothesis beyond the indirect evidence from output variance.

\section{Experiments}

\subsection{Setup}

All experiments use 20 random seeds. We compare conditions across two environment variants (standard and non-stationary) and two training horizons (10{,}000 and 50{,}000 steps). Conditions include:

\begin{itemize}
    \item \textbf{Full (add-on)}: all three self-monitoring modules as auxiliary-loss add-ons ($\sim$37{,}400 parameters)
    \item \textbf{Structural}: all three modules with structural integration ($\sim$38{,}400 parameters)
    \item \textbf{No self-monitoring}: all modules removed ($\sim$26{,}700 parameters)
    \item \textbf{Param-matched}: hidden dimension increased to 40 (38{,}649 parameters), no modules
    \item \textbf{Aux control}: all modules present with matched parameters, but trained against random targets (isolates whether \emph{targeted} vs \emph{any} auxiliary loss matters)
    \item \textbf{Single $\tau$}: all $\tau$ values fixed at 30 (no multi-timescale processing)
\end{itemize}

We use the food-to-death ratio as our primary metric: total food items eaten divided by $\max(\text{total deaths}, 1)$ over the full training run for each seed. In practice, no seed achieved zero deaths. All error bars are sample standard deviations ($\text{ddof}{=}1$). Statistical tests are Welch's $t$-tests; paired $t$-tests are used where the same seeds appear in both conditions. We report Cohen's $d$ effect sizes and note that with $n{=}20$, individual comparisons have moderate power.

\subsection{Phase 1: Add-On Modules Do Not Help}

Table~\ref{tab:results} shows the core null result. In the standard environment with 10{,}000 steps, the full add-on model (food/death ratio $0.75 \pm 0.32$) is statistically indistinguishable from the no-self-monitoring baseline ($0.74 \pm 0.22$), the parameter-matched control ($0.86 \pm 0.53$), and the random-auxiliary control ($0.85 \pm 0.36$). No pairwise comparison reaches significance (all $p > 0.15$).

This null result holds across all tested conditions:

\begin{table}[h]
\centering
\caption{Food/death ratio across conditions (mean $\pm$ std, $n{=}20$ seeds). No condition significantly outperforms the no-self-monitoring baseline in the standard environment. Dashes indicate conditions not run: structural integration was developed after the initial null result and tested only at 50k steps; Single $\tau$ was a diagnostic ablation run only in the standard 10k condition.}
\label{tab:results}
\begin{tabular}{lcccc}
\toprule
\textbf{Condition} & \textbf{Std 10k} & \textbf{Std 50k} & \textbf{Nonstat 10k} & \textbf{Nonstat 50k} \\
\midrule
Full (add-on)    & $0.75 \pm 0.32$ & $1.34 \pm 0.82$ & $0.87 \pm 0.18$ & $0.90 \pm 0.20$ \\
Structural       & --- & $1.56 \pm 0.90$ & $0.84 \pm 0.19$ & $\mathbf{1.08 \pm 0.35}$ \\
No self-monitoring & $0.74 \pm 0.22$ & $1.38 \pm 1.02$ & $0.90 \pm 0.28$ & $1.03 \pm 0.32$ \\
Param-matched    & $0.86 \pm 0.53$ & $1.33 \pm 0.94$ & $0.93 \pm 0.32$ & $1.18 \pm 0.40$ \\
Aux control      & $0.85 \pm 0.36$ & $1.21 \pm 0.76$ & $0.91 \pm 0.22$ & $0.98 \pm 0.30$ \\
Single $\tau$    & $0.87 \pm 0.45$ & --- & --- & --- \\
\bottomrule
\end{tabular}
\end{table}

\subsection{Phase 2: Diagnosing the Failure}

\begin{figure}[t]
\centering
\includegraphics[width=0.95\textwidth]{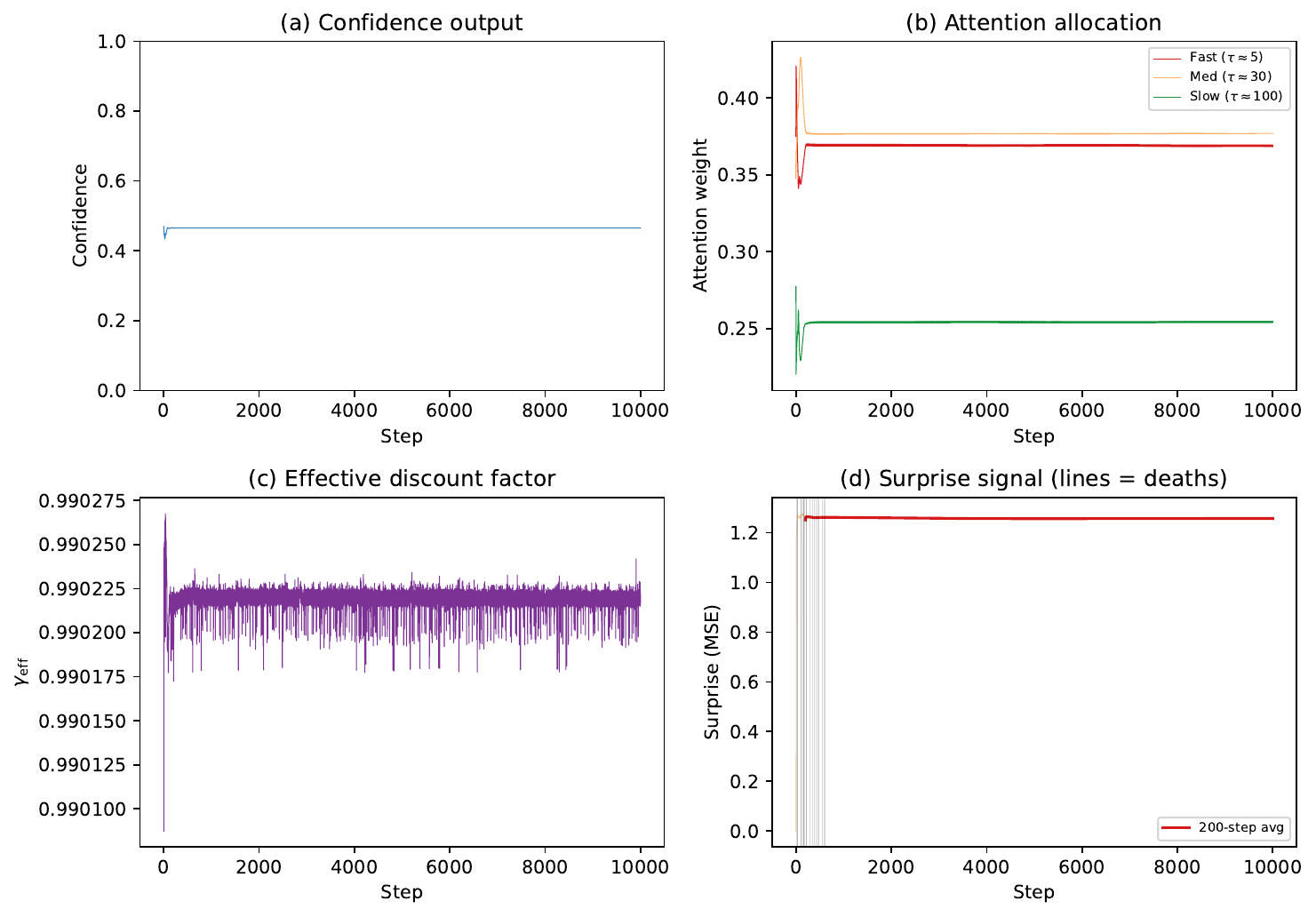}
\caption{Module output collapse (seed 42, add-on design, standard 1D environment). (a)~Confidence is flat at $\sim$0.47. (b)~Attention allocation is constant across cortical levels. (c)~$\gamma_{\text{eff}}$ varies by $< 0.0002$. (d)~Surprise rises from zero (no prediction at step 0) to a plateau within $\sim$200 steps and remains constant; vertical lines mark deaths, which produce no response.}
\label{fig:collapse}
\end{figure}

Why do the modules fail? We examine module outputs in detail for seed 42 (full add-on mode, standard environment) and report summary statistics across all 20 seeds where applicable. Figure~\ref{fig:collapse} visualizes the collapse for seed 42.

\paragraph{Confidence is a constant.} For seed 42, the metacognition module's confidence output has mean $0.47$ and standard deviation $0.002$ across 10{,}000 steps. Across all 20 seeds, the median per-seed confidence standard deviation is $0.0024$ (max $0.0051$). The agent receives a near-constant feature that contributes nothing to decision-making.

\paragraph{Attention allocation does not shift.} The three-way attention allocation over cortical levels has per-level std $< 0.011$ (median across seeds: $0.004$). It does not change around deaths, danger waves, or food encounters. The workspace broadcast modulation by attention allocation is multiplying near-constant weights by near-zero gate values.

\paragraph{Subjective duration has negligible effect.} Felt duration ranges from 0.73 to 0.91 across steps for seed 42. The resulting $\gamma_{\text{eff}}$ ranges from 0.99009 to 0.99027---a change of $0.0002$. The discount factor is effectively constant. The subjective duration module, despite its 273 parameters, does not influence learning.

\paragraph{Policy sensitivity confirms ignore-ability.} To go beyond indirect evidence from output variance, we directly measured policy sensitivity by perturbing each module output and computing $D_{\text{KL}}(\pi_{\text{base}} \| \pi_{\text{perturbed}})$. In the add-on design, perturbing confidence by $\delta = 0.5$ produces mean KL $< 10^{-5}$, confirming that the policy is insensitive to the module outputs. The agent has learned to route decisions through pathways that bypass the module signals entirely.

\paragraph{Root cause: the feedback loop is circular.} The module outputs (meta-signal, self-prediction error, felt duration) are fed back as input features alongside the observation. But these signals are derived from the hidden states, which are themselves derived from the observations. In a fully observable environment, the agent already has all the information it needs. The module outputs are a lossy, delayed re-encoding of data the agent already possesses. There is no inductive bias forcing the agent to \emph{use} these signals, so gradient descent learns to ignore them.

\subsection{Phase 3: Structural Integration}

Table~\ref{tab:results} includes the structural integration condition. The key comparison is in the non-stationary environment with 50{,}000 training steps:

\paragraph{Structural vs add-on.} The structurally integrated agent ($1.08 \pm 0.35$) outperforms the original add-on approach ($0.90 \pm 0.20$) with Cohen's $d = 0.62$ (medium-large effect) and $p = 0.06$ (paired $t$-test). The structural agent wins on 13 of 20 seeds. Since the only change is how module outputs enter the decision pathway, the improvement comes from eliminating the ignore-ability of the add-on design. This result is marginally significant; a larger sample may resolve it.

\paragraph{Add-on vs no self-monitoring.} The add-on agent ($0.90$) trends below the no-self-monitoring baseline ($1.03$) with $d = 0.49$ and $p = 0.14$. This difference is not significant, but the direction is consistent with the hypothesis that ignored modules cause mild harm through gradient competition. We rely on this trend, not a significant result, when we describe the add-on design as harmful below.

\paragraph{Structural vs no self-monitoring.} The structural agent ($1.08$) is not significantly better than the no-self-monitoring baseline ($1.03$, $p = 0.67$, $d = 0.15$). While it trends higher and wins 11/20 seeds, the effect is small. Structural integration recovers from the trend-level harm of the add-on design but does not demonstrate that self-monitoring provides a decisive advantage over having none at all.

\paragraph{Structural vs aux control.} In the standard 50k environment, structural ($1.56$) outperforms random-auxiliary ($1.21$) with $d = 0.42$ and $p = 0.13$. Targeted self-monitoring signals may provide more value than random auxiliary losses, though the difference does not reach significance.

\paragraph{Param-matched comparison.} The param-matched control ($1.18 \pm 0.40$) numerically outperforms the structural agent ($1.08 \pm 0.35$) in the non-stationary 50k condition, though the difference is not significant ($d = 0.27$, $p = 0.40$). Simply increasing hidden dimension from 32 to 40 without any self-monitoring modules produces comparable or better results. This means we cannot rule out that additional capacity, rather than self-monitoring content, drives the structural agent's improvement over the add-on design.

\subsection{Phase 4: Component-Wise Structural Ablation}

To determine which structural pathway contributes most, we tested each pathway in isolation (Table~\ref{tab:component_ablation}). In the non-stationary environment with 50{,}000 steps:

\begin{table}[h]
\centering
\caption{Component-wise structural ablation (non-stationary, 50k steps, $n{=}20$ seeds). Each single-pathway variant activates only one structural integration mechanism.}
\label{tab:component_ablation}
\begin{tabular}{lccc}
\toprule
\textbf{Condition} & \textbf{Food/Death} & \textbf{$d$ vs full structural} & \textbf{$p$} \\
\midrule
Full structural     & $1.08 \pm 0.35$ & --- & --- \\
TSM only            & $1.04 \pm 0.30$ & $0.12$ & $0.70$ \\
Confidence only     & $0.96 \pm 0.25$ & $0.39$ & $0.22$ \\
Surprise only       & $0.94 \pm 0.28$ & $0.44$ & $0.17$ \\
Full (add-on)       & $0.90 \pm 0.20$ & $0.62$ & $0.06$ \\
No self-monitoring  & $1.03 \pm 0.32$ & $0.15$ & $0.67$ \\
\bottomrule
\end{tabular}
\end{table}

The TSM-only variant performs closest to the full structural agent ($d = 0.12$), suggesting that routing self-model predictions into the policy head is the most impactful structural change. The confidence-gating and surprise-triggered workspace pathways each provide a partial improvement over the add-on baseline but do not recover the full structural benefit alone. The three pathways appear complementary, with the TSM-to-policy route carrying the largest individual contribution.

\subsection{Phase 5: 2D Partially Observable Environment}

We also evaluated the add-on modules in a 2D toroidal environment (Section~3.8). Performance is much lower across all conditions (food/death ratios $< 0.06$), indicating that agents barely learn in this harder setting within 10{,}000 steps. The null result for add-on modules holds: the full add-on model ($0.041 \pm 0.015$) does not outperform the no-self-monitoring baseline ($0.057 \pm 0.019$). However, we did not test structural integration in 2D, so we cannot assess whether the architectural lesson transfers. The 2D environment may require longer training horizons or architectural changes to support meaningful learning.

\subsection{Learned Properties}

Despite the null performance result, the self-monitoring modules learn interpretable internal structure:

\paragraph{Predictability hierarchy.} The temporal self-model learns that slow cortical levels ($\tau \approx 100$) are more self-predictable than fast levels ($\tau \approx 5$). This ordering follows from the architecture---slow levels change less per step---but confirms that the self-prediction objective captures meaningful variation across timescales.

\paragraph{Decreasing self-prediction error.} The metacognition module's surprise and the TSM's prediction error both decrease over training. Because these are auxiliary losses, a decrease is expected from optimization. However, the decrease accelerates during periods when survival performance also improves, suggesting the self-prediction and task objectives reinforce each other.

\section{Discussion}

\paragraph{Why add-on self-monitoring fails.}
Self-monitoring modules implemented as auxiliary-loss add-ons collapse to near-constant outputs that the agent learns to ignore. The modules sit outside the decision-making pathway. The agent receives module outputs as input features, but there is no inductive bias requiring their use. Gradient descent finds it easier to ignore these noisy, redundant features than to learn how to exploit them. Policy sensitivity analysis confirms this directly: perturbing module outputs in the add-on design produces negligible KL divergence in the policy distribution. The auxiliary losses still flow gradients through the shared cortical cells, and these gradients may push representations toward being good for self-prediction rather than good for action selection. The trend-level performance drop of the add-on agent relative to the no-self-monitoring baseline in the non-stationary environment ($d = 0.49$, $p = 0.14$) is consistent with such gradient competition, though the effect is not significant and does not appear in the standard environment.

\paragraph{Why structural integration helps.}
Structural integration places module outputs on the critical path, addressing the ignore-ability problem directly. Confidence directly scales the entropy bonus, so the agent's exploration rate depends on it. Surprise triggers workspace broadcasts, so cross-level communication requires it. Self-model predictions enter the policy head, so the agent sees predicted future states alongside current states. The modules are no longer optional input features---they are load-bearing parts of the architecture. Component-wise ablation reveals that the TSM-to-policy pathway carries the largest individual contribution, though the three pathways appear complementary.

The result is a medium-large improvement ($d = 0.62$) over the add-on design in non-stationary environments. However, structural integration does not significantly outperform having no self-monitoring at all ($d = 0.15$, $p = 0.67$). The gain over the add-on design appears to come from preventing the trend-level harm of ignored modules (add-on vs.\ no-self-monitoring: $d = 0.49$, $p = 0.14$), not from self-monitoring per se.

\paragraph{Capacity confound.}
The structural integration changes are not pure re-routings of existing signals: the TSM-to-policy pathway doubles the policy head's input dimension from $d_h$ to $2 d_h$, and the param-matched control (which simply increases hidden dimension without any self-monitoring) numerically outperforms the structural agent in the non-stationary 50k condition. We cannot fully disentangle whether the structural agent's improvement over the add-on design comes from self-monitoring content or from additional representational capacity in the policy head. A control condition providing random features of matching dimension to the policy head would resolve this but is beyond our current experiments.

\paragraph{The auxiliary-loss trap.}
For consciousness-inspired and self-monitoring AI architectures more broadly, our results indicate that auxiliary losses alone are not sufficient to make self-monitoring useful, at least in the environments we tested. Consciousness-inspired architectures~\citep{goyal2022coordination, juliani2022deep} and theoretical accounts of machine metacognition~\citep{cleeremans2020learning} typically frame self-monitoring as something to be trained via auxiliary objectives. Our results show that this approach can produce modules that train successfully (loss decreases, representations form) but have zero impact on behavior. The module learns to predict, but the agent does not learn to \emph{use} the predictions. The relevant design criterion is whether the agent's decision pathway depends on the module outputs---trainability alone is insufficient. This conclusion is drawn from a specific class of environments; the auxiliary-loss approach may prove more effective with richer partial observability or larger model capacity.

\paragraph{Implications for consciousness-inspired AI.}
Theories of consciousness propose self-monitoring mechanisms~\citep{baars1988cognitive, rosenthal2006consciousness, graziano2013consciousness} that are architecturally central to biological cognition, not peripheral add-ons. Our results align with this perspective: self-monitoring is more effective when it sits on the decision pathway. Computational implementations of consciousness theories should attend to \emph{where} self-monitoring sits in the processing architecture---placement matters more than presence.

\paragraph{Limitations.}
Our evaluation spans two environment families (1D and 2D toroidal predator-prey worlds) with non-stationary variants, but both are relatively simple compared to the settings where self-monitoring is most likely to matter. The parameter budget is small ($\sim$37{,}400 total). Training is relatively short (50{,}000 steps maximum). The structural integration variant, while showing improvement over add-on modules, does not significantly outperform the no-self-monitoring baseline or the param-matched control, leaving open whether the modules provide genuine value or whether the architectural changes simply add capacity and reduce the trend-level harm of the add-on design.

The component-wise ablation bundles each structural pathway with its associated module; a fully factorial design crossing module presence with integration pathway, and a control providing random features of matching dimension to the policy head (to resolve the capacity confound discussed above), would provide stronger causal evidence but are beyond our current scope. Policy sensitivity analysis uses a fixed perturbation magnitude; dose-response curves across multiple magnitudes would strengthen the findings.

Larger environments with richer dynamics---multi-agent settings, partially observable 3D worlds, hierarchical task structures, continual learning under distribution shift---might reveal clearer benefits of self-monitoring and are important directions for future work.

The random-auxiliary control tests whether \emph{any} auxiliary loss helps, but it does not test every possible alternative. Specific non-random auxiliary tasks (e.g., reward prediction, next-action prediction) might provide comparable benefits without self-monitoring. Our param-matched control and aux-control together bound the contribution, but comparing against every possible auxiliary objective is beyond scope.

With $n{=}20$ seeds, our statistical power is moderate. The structural-vs-add-on comparison ($p = 0.06$) is borderline; a larger sample might resolve it. We report effect sizes throughout to complement significance tests.

\paragraph{Representation without function.}
The self-monitoring modules learn interpretable internal structure (a predictability hierarchy, decreasing self-prediction error) even when they do not improve task performance. Whether these learned representations constitute anything resembling self-knowledge is a question behavioral data alone cannot answer~\citep{nagel1974like}. What we can say is that the \emph{functional} value of self-monitoring depends on architectural integration.

\section{Conclusion}

We investigated whether self-monitoring modules---metacognition, temporal self-models, and subjective duration---improve continuous-time multi-timescale RL agents. The answer depends on \emph{how} the modules are integrated.

As auxiliary-loss add-ons, the modules provide no statistically significant benefit across 20 seeds in both 1D and 2D predator-prey environments, and training up to 50{,}000 steps. Diagnosis reveals their outputs collapse to near-constants that the agent ignores, and policy sensitivity analysis confirms the agent's decisions are unaffected by these signals.

Structurally integrating the module outputs---using confidence for exploration gating, surprise for workspace triggering, and self-model predictions as policy inputs---improves over the add-on design (Cohen's $d = 0.62$) in non-stationary environments. Component-wise ablation shows the TSM-to-policy pathway contributes most to this gain. However, structurally integrated self-monitoring does not significantly surpass a baseline with no self-monitoring ($d = 0.15$, $p = 0.67$), and a param-matched control without modules performs comparably. The primary benefit of structural integration appears to be recovering from the trend-level harm of the add-on design rather than extracting value from self-monitoring per se.

The practical lesson is that self-monitoring modules are more effective on the decision pathway than beside it. Auxiliary losses can train modules that learn interesting internal representations, but unless the agent's action selection depends on the module outputs, the modules are behaviorally inert. Whether self-monitoring can provide a clear advantage over no self-monitoring likely depends on environment complexity and partial observability.

Future work should explore settings where self-monitoring should matter more strongly---partially observable multi-agent environments, continual learning under distribution shift, or hierarchical planning---and should investigate whether the structural integration principle holds at larger model scales. Component-wise factorial designs crossing module presence with integration pathway, a random-feature control to isolate the capacity confound in the TSM-to-policy pathway, and dose-response sensitivity analysis across perturbation magnitudes, would provide stronger causal evidence for the mechanisms at work.

\section*{Acknowledgments}
Claude Code (Anthropic) was used to assist in implementing the experimental code.

\bibliographystyle{plainnat}

\end{document}